\documentclass[twoside]{article}

\usepackage[accepted]{aistats2025}

\usepackage{hyperref}       % hyperlinks
\usepackage{url}
\usepackage{amsmath}
\usepackage{amssymb}
\usepackage{mathtools}
\usepackage{amsthm}
\usepackage{algorithm}
\usepackage{algorithmic}
\usepackage{booktabs}
\usepackage{float}

\newtheorem{theorem}{Theorem}
\newcommand{\innerproduct}[2]{\langle #1, #2 \rangle}

\begin{document}

\twocolumn[

\aistatstitle{A Trust-Region Method for Graphical Stein Variational Inference}

\aistatsauthor{Liam Pavlovic \And David M. Rosen}

\aistatsaddress{Northeastern University}]

\begin{abstract}
  Stein variational inference (SVI) is a sample-based approximate Bayesian inference technique that generates a sample set by jointly optimizing the samples’ locations to minimize an information-theoretic measure of discrepancy with the target probability distribution. SVI thus provides a fast and significantly more sample-efficient approach to Bayesian inference than traditional (random-sampling-based) alternatives. However, the optimization techniques employed in existing SVI methods struggle to address problems in which the target distribution is high-dimensional, poorly-conditioned, or non-convex, which severely limits the range of their practical applicability. In this paper, we propose a novel trust-region optimization approach for SVI that successfully addresses each of these challenges. Our method builds upon prior work in SVI by leveraging conditional independences in the target distribution (to achieve high-dimensional scaling) and second-order information (to address poor conditioning), while additionally providing an effective adaptive step control procedure, which is essential for ensuring convergence on challenging non-convex optimization problems. Experimental results show our method achieves superior numerical performance, both in convergence rate and sample accuracy, and scales better in high-dimensional distributions, than previous SVI techniques.
\end{abstract}

\section{Introduction}

Drawing inferences from noisy data is a fundamental capability in artificial intelligence, machine learning, and scientific and engineering applications. Mathematically, this procedure is naturally expressed in the language of posterior Bayesian inference. Many of these inference problems can be formulated as probabilistic graphical models (PGMs), which are an effective tool for modeling joint distributions with known conditional independences among the individual variables. The conditional independence structure encoded in a PGM can greatly simplify the inference task \cite{pgm}. Nonetheless, exact Bayesian inference is typically computationally intractable, so in practice approximate inference methods are used instead.

One of the most common approximate Bayesian inference methods is sample-based approximation, which uses a sample set to represent the target distribution. This approximation has the benefits of simplicity, flexibility, arbitrary precision (as sample size increases), and easy empirical estimation of any statistic over the target distribution. Traditional methods for generating a sample-based approximation are based on random sampling. Common examples of random sampling algorithms include Markov chain Monte Carlo (MCMC) \cite{mcmc}, nested sampling \cite{dynesty}, and Hamiltonian Monte Carlo \cite{hmc}. Due to their dependence on random processes to explore the state space, these methods can be slow to converge and sample inefficient. These deficiencies become especially pronounced in high-dimensional problems. 

Stein variational inference (SVI) is a more efficient alternative for generating a sample-based approximation \cite{svgd}. In place of random sampling, SVI deterministically optimizes a set number of samples to minimize KL divergence with the reference distribution. SVI has been demonstrated to have superior sample efficiency over random sampling methods, capturing more information with fewer samples \cite{svgd}. However, SVI can still struggle to scale to high-dimensional, ill-conditioned, and non-convex objectives. Previous work \cite{mpsvgd, graphsvi, svn} on SVI methods has addressed some of these challenges individually, but no single previous SVI method for PGM problems handles all these challenges well.

In this paper, we propose a novel trust-region optimization approach for SVI that successfully addresses each of these challenges. Our method builds upon prior work by leveraging both conditional independences in the target distribution to achieve high-dimensional scaling \cite{mpsvgd, graphsvi} and second-order information to address poor conditioning \cite{svn} in the same system. We also provide an effective adaptive step control procedure for SVI, which is essential for ensuring convergence on challenging non-convex optimization problems. Experimental results show our method achieves superior numerical performance, both in convergence rate and sample accuracy, and scales better to high-dimensional distributions than previous variational inference techniques.

\section{Stein Variational Inference}
The objective of SVI \cite{svgd} is to approximate a given target distribution $p(x)$ on $\mathcal{X} \subseteq \mathbb{R}^D$ using the Kullback-Leibler (KL) divergence-minimizing representative $q$ within some family $Q$ of tractable model distributions
\begin{equation} \label{svi-objective}
    \min_{q \in Q} KL(q || p) \equiv \mathbb{E}_{x \sim q} [\log q(x) - \log p(x)]
\end{equation}
To achieve this, we begin with some initial distribution $q_0$ and generate a set of pushforwards $q_1, ..., q_L$ according to the rule $q_{l+1} = (T_l)_*q_l$ where $T_l \in \mathbb{R}^D \rightarrow \mathbb{R}^D$ is some perturbation $T_l = I + \Phi_l$ of the identity map. At each iteration $l$, we seek a perturbation function $\Phi_l$ from some function space $\mathcal{F}$ such that 
\begin{equation} \label{svi-inner-opt}
    J[\Phi_l] \triangleq KL((I+\Phi_l)_*q_l||p) \quad J[\Phi_l] < J[\mathbf{0}]
\end{equation}
To ensure the descent condition, we can choose $\Phi_l$ to be an infinitesimal application of the negative functional gradient at $J[\mathbf{0}]$ in $\mathcal{F}$. For a Hilbert space $\mathcal{F}$, the gradient of a functional $J$ at some $S \in \mathcal{F}$ is defined as the element $\nabla J[S] \in \mathcal{F}$ satisfying 
\begin{equation*}
\innerproduct{\nabla J[S]}{V}_\mathcal{F} = DJ[S](V) \quad \text{for all } V \in F
\end{equation*}
\begin{equation*}
 DJ[S](V) \triangleq \lim_{\tau \rightarrow 0} \frac{1}{\tau}(J[S + \tau V] - J[S]) 
\end{equation*}

Consider some kernel $k: \mathcal{X} \times \mathcal{X} \mapsto \mathbb{R}$ with a corresponding reproducing kernel Hilbert space (RKHS) $\mathcal{H}$. A particularly advantageous choice for $\mathcal{F}$ is the vector-valued RKHS $\mathcal{H}^D$ because this space has a closed form for the desired functional gradient $\nabla J[\mathbf{0}]$ \cite{svgd}
\begin{multline}\label{phi-max}
    \nabla J[\mathbf{0}](x) = -\mathbb{E}_{z\sim q}[k(z, x)\nabla_z \log p(z) \\ 
    + \nabla_z k(z, x)]
\end{multline}

In order to compute the expectation above, we need some tractable representation of the distribution $q$. A natural choice for this representation is a \textit{sample}, since this parameterization is both flexible, and makes the expectation in Eq. \ref{phi-max} trivial to approximate. 

Stein variational gradient descent (SVGD) \cite{svgd} combines a sample-based approximation of $q$ and the descent direction in Eq. \ref{phi-max} into an iterative procedure for generating an sample-based approximation of $p$. SVGD first samples a set of points $\{x_i\}_{i=1}^n$ from an initial distribution $q_0$ and then iteratively updates the location of each sample using some static step size $\xi$ and a sample-based approximation of Eq. \ref{phi-max}
\begin{multline}
    x_i \leftarrow x_i + \xi\frac{1}{n}\sum_{j=1}^n k(x_j, x_i) \nabla_{x_j} \log p(x_j) \\
    + \nabla_{x_j}k(x_j, x_i)
\end{multline}

The first term of this update pushes the samples towards high probability areas of $p$ while the second term, referred to as the \textit{kernel repulsion term}, pushes apart samples that are close together. The kernel repulsion term is essential because it spreads the sample across the distribution \cite{svgd}.

\section{Related Work on SVI}
Since this paper is primarily concerned with improving SVI specifically, we restrict our discussion of related work to SVI methods. SVGD \cite{svgd}, as discussed above, struggles to scale to high-dimensional, non-convex, and ill-conditioned objectives. Several subsequent works have suggested modifications to address these challenges.

There are two major contributors to SVI's poor performance in high dimensions. First, the information (in bits) required to encode the joint target distribution $p$ grows exponentially with respect to it's dimension. This increases the amount of information approximation methods (like SVI) must infer to accurately approximate $p$ \cite{pgm}. Second, when using distance-based kernels, like the radial basis function (RBF) kernel, the magnitude of the kernel repulsion term decreases in higher dimensions, resulting in mode collapse, where all the sample points are densely packed around a single mode of the target distribution \cite{mpsvgd}. When the conditional factorization of $p$ is known (i.e. when $p$ is represented by a PGM), these conditional independence relations can be exploited to dramatically simplify the inference task and address these challenges \cite{pgm}. 

Graphical Stein variational inference methods \cite{mpsvgd, graphsvi} exploit the conditional independences encoded in the PGM for the target $p$ via their kernel design. Assuming the space $\mathcal{X}$ is a product of factors $\mathcal{X} = \mathcal{C}_1 \times ... \times \mathcal{C}_D$, graphical SVI employs a set of $D$ local kernels $k_a : \mathcal{X}_{\mathcal{S}_a} \times \mathcal{X}_{\mathcal{S}_a} \mapsto \mathbb{R}$, where $\mathcal{S}_a$ represents the factor $\mathcal{C}_a$ and its Markov blanket $\Gamma_a$. In this local kernel setting, the Hilbert space $\mathcal{F}$ over which we take functional gradients becomes the product $\mathcal{H}_1 \times ...\times\mathcal{H}_D$ of the local kernels' RKHSs. In this space, the closed form for the functional gradient $\nabla \hat{J}[\mathbf{0}]$ ($\wedge$ decorator used to indicate usage of local kernels and differentiate from the gradient in Eq. \ref{phi-max}) is given by 
\begin{multline}\label{mp-grad}
(\nabla \hat{J}[\mathbf{0}](x))_a =-\mathbb{E}_{z \sim q} [k_a(z, x) \nabla_{z_a} \log p(z) \\
+ \nabla_{z_a}k_a(z, x)] \quad \text{for } a \in 1, ..., D
\end{multline}
This functional gradient interpretation of the graphical SVI update follows naturally from previous work \cite{mpsvgd, graphsvi}, but has not been presented before. So, we present a proof in Appendix \ref{mp-grad-proof}. Note that utilizing the above descent direction only guarantees that the resulting approximation $q$ agrees with the target's \textit{conditional} distributions: $q(x_a | x_{\Gamma_a}) = p(x_a | x_{\Gamma_a})$ \cite{mpsvgd, graphsvi}, indicating that these methods inference the conditionals as desired. 

Other work on SVI has explored improving high-dimensional performance on problems without conditional structure by employing Grassman manifold \cite{grassmann}, low-dimensional subspace projection \cite{psvgd, psvn}, and slicing \cite{ssvgd} strategies. 

 Many previous SVI methods only utilize first-order information in their updates, which is often insufficient to achieve good convergence on ill-conditioned objectives. The Stein variational Newton method (SVN) \cite{svn} incorporates second-order information by deriving a Newton system to compute an approximate Newton update $w_i$ for each sample $x_i$. This Newton system is block-structured
\begin{equation*} 
    \begin{bmatrix}
    H(x_1, x_1) & \dots  & H(x_1, x_n)\\
    \vdots & \ddots & \vdots\\
    H(x_n, x_1) & \dots  & H(x_n, x_n) 
    \end{bmatrix}    
    \begin{bmatrix}
    w_1\\
    \vdots \\
    w_n\\
    \end{bmatrix}=
    \begin{bmatrix}
    -\nabla J[\mathbf{0}](x_1)\\
    \vdots \\
    -\nabla J[\mathbf{0}](x_n)\\
    \end{bmatrix}
\end{equation*}
where the $ab$-th entry of the Hessian matrix block $H(x, y)$ is
\begin{multline}
    (H(x, y))_{ab} = 
     \mathbb{E}_{z \sim q} [-k(z, x)k(z, y)\partial_{ab} \log p(z) \\
     + \partial_{z_b}k(z, x)\partial_{z_a}k(z, y)]
\end{multline}

By applying a block-diagonal approximation to the original system, SVN solves a decoupled system for the approximate Newton update $w_i$ of each sample $x_i$
\begin{equation} \label{svn-system}
    H(x_i, x_i) w_i = -\nabla J[\mathbf{0}](x_i) \quad \text{for } i \in 1, ..., n
\end{equation}

 To solve these decoupled systems, SVN utilizes a conjugate gradient method such as the Newton-CG method \cite{nocedal}. Like SVGD, SVN does not leverage conditional independence, and thus performs poorly in high dimensions. 

Importantly, the previously discussed graphical SVI methods do not utilize an adaptive method for step control, which was identified in \cite{svn} as important future work. Adaptive step control has been implemented for low-dimensional subspace projection methods \cite{psvn}. However, this method selects step sizes to minimize the KL divergence between \textit{projections} of the target $p$ and sample set $\{x_i\}_{i=1}^n$ onto a low-dimensional subspace, which does not necessarily guarantee a reduction of the divergence between the target $p$ and sample set $\{x_i\}_{i=1}^n$ themselves.   

%Other work on direct sample optimization \cite{ksdd} has implemented step control a different, more tractable objective than SVI but did not outperform vanilla SVGD \cite{svgd}. 

\section{Exploiting Conditional Independence in Second-Order SVI}
\subsection{Second Variation in Local Kernel Setting}
As a first contribution, we show how to implement a second-order Hessian model for SVI that exploits conditional independence. To do this, we generalize the formula for the second-order variation presented in SVN \cite{svn} to the local kernel space $\mathcal{H}_1 \times ... \times \mathcal{H}_D$ utilized by graphical SVI methods. The second variation is defined as the directional derivative along a pair of directions $V, W \in \mathcal{H}_1 \times ... \times \mathcal{H}_D$ 
\begin{equation*}
    D^2J[\mathbf{0}](V, W) = \lim_{\tau \rightarrow 0} \frac{1}{\tau}(DJ[\tau W](V) - DJ[\mathbf{0}](V))
\end{equation*}
\begin{theorem} \label{second-var-theorem}
    Along a pair of directions $V, W \in \mathcal{H}_1 \times ... \times \mathcal{H}_D$ the second variation is \footnote{The inner products here are between the functions $h_{ab}$, $w_b, v_a$ in Hilbert spaces. $x$ and $y$ are free variables only included to show which functions share which inputs.}
    \begin{multline}
         D^2J[\mathbf{0}](V, W) = \\
         \sum_{a=1}^D \sum_{b=1}^D\innerproduct{\innerproduct{h_{ab}(x, y)}{w_b(y)}_{\mathcal{H}_b}}{v_a(x)}_{\mathcal{H}_a}
    \end{multline}
    \begin{multline}\label{mp-hessian}
        h_{ab}(x, y) = \mathbb{E}_{z \sim q} [-k_a(z, x)k_b(z, y)\partial_{ab} \log p(z) \\
        + \partial_{z_a}k_b(z, y)\partial_{z_b}k_a(z, x)]
    \end{multline}
\end{theorem}
For the proof of this theorem, see Appendix \ref{second-var-proof}. 

\subsection{Decoupled Newton Systems}
Following SVN \cite{svn}, we employ a block-diagonal approximation of the Hessian defined by Theorem \ref{second-var-theorem} and solve a decoupled system for the approximate Newton update $w_i$ for each sample $x_i$
\begin{equation} \label{tr-svi-system}
    \hat{H}(x_i, x_i) w_i = -\nabla \hat{J} [\mathbf{0}](x_i) \quad \text{for } i \in 1, ..., n
\end{equation} 
     
where the entries of the Hessian $\hat{H}(x_i, x_i)$ are defined by the second variation coefficients $(\hat{H}(x_i, x_i))_{ab} = h_{ab}(x_i, x_i)$ from Eq. \ref{mp-hessian}.

\section{Trust-Region Methods}
Our second contribution is implementing two trust-region methods for SVI optimization. Trust-region methods are iterative procedures for optimizing a smooth objective function $f: \mathbb{R}^M \rightarrow \mathbb{R}$ \cite{nocedal}. At each iteration $t$, these methods generate an additive update $w \in \mathbb{R}^M$ to the current estimate $x \in \mathbb{R}^M$ by minimizing a second-order approximation of the objective over a closed ball defined by some norm $||\cdot||$, called the \textit{trust region}
\begin{multline} \label{trust-region-eq}
\text{min}_{w \in \mathbb{R}^M} f(x) + \nabla f(x)^\top w + \frac{1}{2}w^\top H w \\ 
\text{s.t. } ||w|| \leq \Delta
\end{multline}
where $H$ is the symmetric Hessian model.

Given the Newton system's (Eq. \ref{tr-svi-system}) block-diagonal structure, we propose utilizing the norm $||w|| \triangleq \max\{||w_i||_2\}_{i=1}^n$ to define the trust region, where $w_i$ is the update for each individual sample $x_i$. The advantage of using this norm is that, in combination with the block-diagonal Hessian model, the trust-region minimization Eq. \ref{trust-region-eq} is \textit{separable} over the updates $w_i$ for each sample $x_i$; consequently, these updates can be efficiently computed in parallel.

\subsection{KL Divergence Approximation}

Standard trust-region methods rely on objective function evaluations to iteratively adjust the trust-region radius $\Delta$ used in each iteration based on the observed change in objective value. In the specific case of SVI, we aim to minimize KL divergence. The computation of KL divergence can be split into two terms
\begin{equation*}
    KL(q||p) = \mathbb{E}_{x \sim q}\left[ -\log p(x) \right] - \mathcal{H}(q)
\end{equation*}
where $\mathcal{H}(q)$ is the entropy of $q$. The first term can be easily approximated via an empirical estimate with our current sample set. The second term requires computing the entropy of $q$ given the representative set of samples $\{x_i\}_{i=1}^n$, which is more difficult. Since SVI already requires the computation of kernel matrices, we propose utilizing the kernelized approximation of entropy from \cite{bach}. 
\begin{equation}
    \mathcal{H}(q) \approx -\mathrm{tr}\left[\frac{1}{n} K \log(\frac{1}{n}K)\right] = -\sum_{i=1}^n \lambda_i \log(\lambda_i)
\end{equation}
where $K$ is the $n \times n$ kernel matrix for the sample $\{x_i\}_{i=1}^n$, such that $K_{ij} = k(x_i, x_j)$, and $\{ \lambda_i\}_{i=1}^n$ are the eigenvalues of $\frac{1}{n}K$. Since this approximation requires the computation of eigenvalues, which is a computationally expensive $O(n^3)$ operation, we utilize the eigenvalues of a Nystr\"om approximation of $\frac{1}{n}K$ in place of the full matrix. We use a non-local kernel $k: \mathcal{X} \times \mathcal{X} \rightarrow \mathbb{R}$ for this approximation. The full algorithmic details of the approximation are presented in Algorithm \ref{approx-KL}. It's time complexity is $\mathcal{O}(m^3 + n)$ where $m$ is the Nystr\"om size.
\begin{algorithm}[h!]
   \caption{Approx-KL}
   \label{approx-KL}
\begin{algorithmic}[1]

    \STATE {\bfseries Inputs:} Sample points $\{x_i\}_{i=1}^n$, reference distribution $p$, Nystr\"om size $m$
    
    \STATE Select a subset $S \subset \{x_i\}_{i=1}^n$ of size $m$ uniformly at random without replacement
    \STATE Compute the kernel matrix $K$ using kernel function $k$ on the subset $S$
    \STATE $U, \{ \lambda_i\}_{i=1}^m, V= \mathrm{SVD}(\frac{1}{n}K)$
    \STATE $H = \sum_{i=1}^m \lambda_i \log(\lambda_i)$
    \STATE $P = \frac{1}{n}\sum_{i=1}^n \log p(x_i)$
    \STATE \textbf{Return} $-P + H$
\end{algorithmic}
\end{algorithm}

\begin{algorithm}[tb]
   \caption{TR-SVI-KL}
   \label{tr-svi-kl}
\begin{algorithmic}[1]
\STATE {\bfseries Inputs:} Initial points $\{x_i\}_{i=1}^n$, reference distribution $p$, initial trust-region $\Delta$

\FOR{each iteration $t$}
    \FOR{each sample $x_i$}
        \STATE Compute $w_i$ by solving system in Eq. \ref{tr-svi-system} using CG-Steihaug \cite{nocedal} with trust-region $\Delta$
    \ENDFOR
    \STATE $m =\sum_{i=1}^n \frac{1}{2} w_i^\top \hat{H}(x_i, x_i)w_i  + \nabla \hat{J}[\mathbf{0}](x_i)^\top w_i$
    \STATE $u = \text{Approx-KL}(\{x_i + w_i\}_{i=1}^n, p, \lfloor n/10 \rfloor)$
    \STATE $o = \text{Approx-KL}(\{x_i\}_{i=1}^n, p,  \lfloor n/10 \rfloor)$
    \STATE $\rho = \frac{u-o}{m}$

    \STATE 
        $\Delta =
        \begin{cases}
            \Delta/ 2 & \text{if } \rho < .0001 \\
            1.5\Delta & \text{if } \rho > .7 \\
            \Delta & \text{otherwise}
        \end{cases}$

    \STATE $\{x_i\}_{i=1}^n = 
        \begin{cases}
           \{x_i\}_{i=1}^n & \text{if } \rho < 0 \\
            \{x_i + w_i\}_{i=1}^n & \text{otherwise}
        \end{cases}$   
\ENDFOR
\end{algorithmic}
\end{algorithm}
Our first trust-region algorithm, TR-SVI-KL, adjusts the shared trust-region size based on how well the local quadratic model (Eq. \ref{trust-region-eq}) predicts the observed change in objective value. If a step increases the objective value, it is rejected. The full algorithmic details of this trust-region method are presented in Algorithm \ref{tr-svi-kl}. The per-iteration time complexity of this method is $\mathcal{O}(n^2D^2 + m^3)$ where $m$ is the Nystr\"om size used for Approx-KL.

\subsection{Gradient-Based Trust-Region}
\begin{algorithm}[tb]
   \caption{TR-SVI-AT}
   \label{tr-svi-at}
            
\begin{algorithmic}[1]
   \STATE {\bfseries Inputs:} Initial points $\{x_i\}_{i=1}^n$, reference distribution $p$
   \STATE $b_{min} =.1$
   \STATE $b,w,b_{max},g = \sqrt{\sum_{i=1}^n||\nabla \hat{J}[\mathbf{0}](x_i)||^2}$
   \FOR{each iteration $t$}
        \FOR{each sample point $x_i$}
            \STATE Compute $w_i$ by solving system in Eq. \ref{tr-svi-system} using CG-Steihaug \cite{nocedal} with trust-region $\frac{g}{b}$
        \ENDFOR
        \STATE $\{x_i\}_{i=1}^n = \{x_i + w_i\}_{i=1}^n $
        \STATE $g = \sqrt{\sum_{i=1}^n||\nabla \hat{J}[\mathbf{0}](x_i)||^2}$
        \STATE $b, w = \begin{cases}
            \max(b_{min}, .9b), g & \text{if } g < .999w \\
            \min (b_{max}, b + \frac{g^2}{b}), w & \text{otherwise}
        \end{cases}
        $
        
    \ENDFOR
\end{algorithmic}
\end{algorithm}
Even with the Nystr\"om approximation, the KL divergence approximation in Algorithm \ref{approx-KL} is still relatively expensive to compute. Moreover, the approximation error can negatively impact the efficacy of the trust-region adjustment. Therefore, in this subsection, we describe an alternative trust-region method that avoids the need to evaluate the objective at all, by taking advantage of the recently developed AdaTrust method \cite{adatrust}. 

The motivating idea behind a gradient-based trust-region is that a converging optimization should contain a subset of iterations in which the magnitude of the gradient is consistently decreasing. Our gradient-based trust-region method stores the lowest observed gradient magnitude value and compares the gradient magnitude at each new iterate against it. The trust-region is expanded if the gradient magnitude at the current iterate is less than the lowest seen so far and constricted otherwise. The algorithmic details of this trust-region method are presented in Algorithm \ref{tr-svi-at}. The per-iteration time complexity of this method is the same as SVN, $\mathcal{O}(n^2D^2)$.

\section{Results}

\subsection{Experimental Set-Up} \label{SetUp}
In this section, we experimentally evaluate our trust-region SVI algorithms. As baselines, we compare against prior work on SVI, namely MP-SVGD \cite{mpsvgd} and SVN \cite{svn}, which also serve as ablations of our method. Motivated by the results in Appendix \ref{add-conv-res}, we utilize variants with more advanced step control rules to make these baselines as strong as possible. The first of these modified methods is MP-SVGD-DLR, which utilizes a decaying step size. The second is MP-SVGD-AG which utilizes the off-the-shelf adaptive optimizer AdaGrad \cite{adagrad} for step control. The third is SVN-CTR which utilizes the CG-Steihaug method \cite{nocedal} for solving the systems in Eq. \ref{svn-system} with a constant trust-region size. The hyperparameters of these variants were fit to each specific problem via grid search to maximize performance, which introduces extra information that our methods did not receive. Note that this makes the following comparisons somewhat unfairly biased against our methods. All SVI methods, both ours and the baselines, utilize an RBF kernel with a lengthscale set manually based on performance. Exact hyperparameter settings for each method and problem are shown in Appendix \ref{hyperparams}. We also compare against VIPS40 \cite{vips}, a state-of-the-art, Gaussian mixture model-based variational inference method. All experiments were run on a desktop with an Intel Core i7-13700K, 32 GB RAM, and a NVIDIA RTX 4080. 

\begin{table*}[tb]
  \caption{Maximum Mean Discrepancies against Ground Truth Sample}
  \label{table}
  \centering
  \begin{tabular}{llll}
    \toprule
    Model    & 12-Dimensional SNLP   & 30-Dimensional BN & 80-Dimensional BN\\
    \midrule
    MP-SVGD-DLR & $0.05276 \pm 0.03560$ &  $0.1492 \pm 0.00618$& $0.2251 \pm 0.00577$ \\
    MP-SVGD-AG & $0.05091 \pm .01328$ & $0.2130 \pm 0.02086$ & $0.2004 \pm 0.00975$\\
    SVN-CTR     & $0.05067 \pm 0.00937$ &    $0.1887 \pm 0.00877$  & $0.2707 \pm 0.00276$\\
    VIPS40   & $0.1981 \pm 0.12383$     &  $\mathbf{0.00185 \pm 0.00095}$ & $0.2565 \pm 0.03055$  \\
    \textbf{TR-SVI-KL}  & $0.04800 \pm 0.00979$ & $0.01496 \pm 0.00741$  & $0.08634 \pm 0.02348$\\
    \textbf{TR-SVI-AT}  & $\mathbf{0.03530 \pm  0.01366}$  & $0.009674 \pm 0.00623$  & $\mathbf{0.07646 \pm 0.02051}$\\
    
    \bottomrule
  \end{tabular}
\end{table*}
\subsection{Bayes Net Experiment}

Our first set of experiments is designed to test the different variational inference methods' ability to scale to distributions that are high-dimensional and poorly conditioned in a controlled environment where we can easily recover ground truth samples from the reference distribtuion. To this end, we evaluate performance on recovering the joint density of a Bayes net \cite{pgm}. This synthetic problem gives us a high degree of control over the parameters of the distribution and the generative nature of Bayes nets enables us to easily recover a ground truth sample via ancestral sampling. This problem also has known conditional independence structure that our methods and graphical baselines can exploit.

The nodes of the Bayes net are organized into layers with nodes in each layer conditioned only on nodes from the previous layer. The conditional/marginal distributions encoded by each node are either a Gaussian or Gaussian mixture. Note that this makes the resulting joint of the Bayes net a Gaussian mixture. The exact parameters of each node are generated randomly according to the generative process described in Appendix \ref{bn-generation}. The generative parameters for the Gaussian mixture nodes are selected to encourage distinct modes. The nodes are also generated to have vastly different variances to induce poor-conditioning. We test on two Bayes nets examples, one 30-dimensional and one 80-dimensional. A ground truth sample for each example, containing 6 million points, was generated via ancestral sampling. Each variational inference method was used to produce a sample with 200 points.

We evaluate the quality of each method's sample by comparing against the ground truth sample using the maximum mean discrepancy (MMD) metric \cite{mmd}. MMD was chosen because it is designed to find the test statistic that reveals the greatest discrepancy between two distributions. MMD is also kernel-based which makes it a somewhat natural choice to evaluate SVI methods. The MMD of two samples $X = \{x_i\}_{i=1}^n$ and $Y = \{y_i\}_{i=1}^m$ is computed as
\begin{multline} \label{mmdeq}
        \text{MMD}(X, Y) = \frac{1}{n^2}\sum_{i,j=1}^n k(x_i, x_j) - \frac{2}{nm}\sum_{i=1}^n\sum_{j=1}^m k(x_i, y_j) \\
        +  \frac{1}{m^2}\sum_{i, j=1}^m k(y_i, y_j)
\end{multline}
For our tests, the kernel $k$ is the RBF kernel whose lengthscale is set using the median heuristic \cite{median-heuristic} on the ground truth sample. The performance for each variational inference method on this metric is shown in Table \ref{table}. All numerical results are averaged over 5 runs with randomly initialized positions for the SVI particle sets. The standard deviation of the MMD statistic over these runs is reported in the error bars.

Our methods outperformed the SVI baselines on both Bayes net problems, suggesting better scaling to high dimensions and poor conditioning. VIPS40 did outperform our method on the lower-dimesnional Bayes net problem. That said, since the joint distribution of this Bayes net is a Gaussian mixture, the Gaussian mixture-based approximation employed by VIPS40 is the information-theoretically optimal choice in this specific case. The difference in performance between our methods and VIPS40 on this problem represents the cost of flexibility. Although our methods can more accurately capture a wider variety of distribution shapes, they cannot achieve gold-standard performance of parametric models in their ideal use cases. Furthermore, our method maintains similar performance on both the 30- and 80-dimensional Bayes Net, while VIPS40 performs two orders of magnitude worse on the higher-dimensional problem. We attribute the pronounced decrease in VIPS40's performance to the fact that VIPS40 does not take advantage of the conditional independences present in these Bayes net models, the exploitation of which is well-known to be critical for achieving efficient inference in high dimensions.
\begin{figure*}[t]
  \centering
  \includegraphics[width=.9\linewidth]{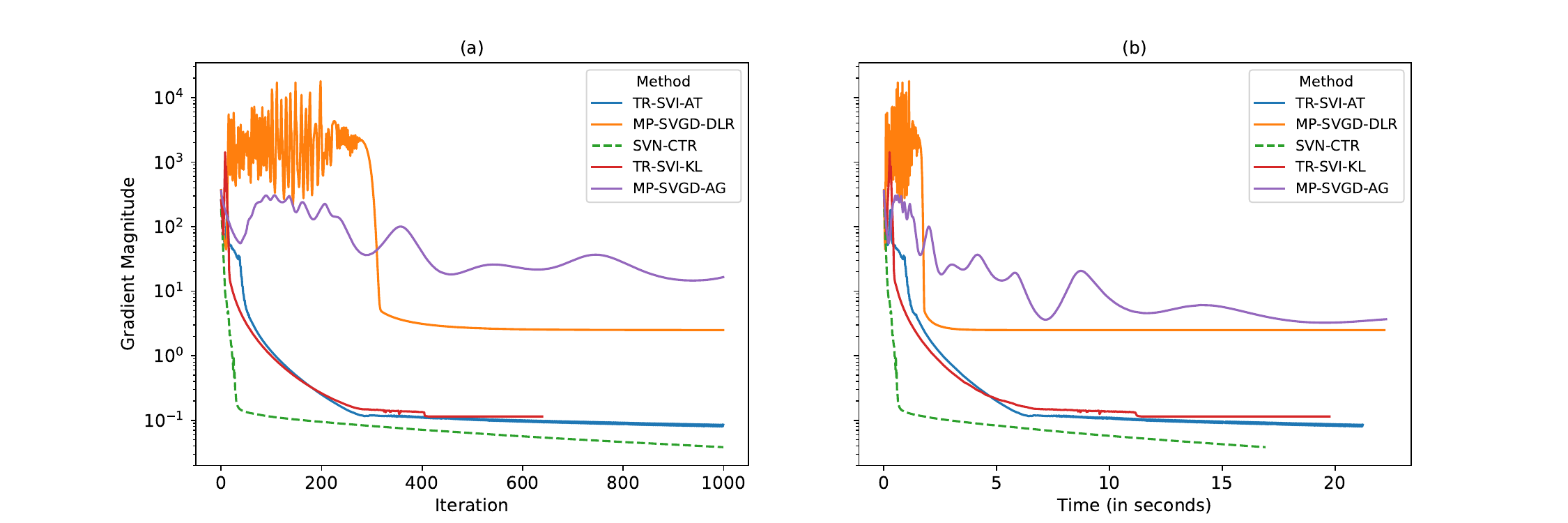}
  \caption{The convergence rate as a function of iteration number (a) and compute time (b) of each SVI method on the small SNLP instance. All second-order methods show fast, smooth convergence both MP-SVGD variants oscillate until their step size decays enough to enable convergence. Note that, unlike the other methods, SVN-CTR does not use local kernels to compute the gradient (see Eqs. \ref{phi-max} and \ref{mp-grad}). Since the estimated gradients depend upon the choice of kernel, SVN-CTR's gradient magnitude values are not directly comparable.}
  \label{converge-best}
\end{figure*}

\subsection{Low-dimensional SNLP with Ground Truth}

Our next set of experiments is designed to highlight the flexibility of our methods by investigating both the convergence behavior and posterior approximation quality of the different variational inference methods on a real-world problem with complex posterior shapes. To this end, we evaluate our methods and the baselines on a small example of the sensor network localization problem (SNLP) \cite{snlp}. This problem tends to have posteriors with multimodal and annular shapes and is therefore hard for parametric models to accurately approximate. Its low dimensionality enables us to recover a ground truth sample for quantitative analysis. 

The goal of the SNLP is to recover the positions of a set of sensors $S = \{s_1, \ldots, s_n\} \subset \mathbb{R}^D$ from a given set of noisy pairwise distance measurements between them.  This problem can be modeled as a graph $G = (V,E)$ in which the sensors are represented by the vertices $V$ and the available measurements $\tilde{d}_{ij}$ are represented by the edges $E$ between them.  A pair of sensors $s_i$ and $s_j$ can only generate a range measurement when the distance $||s_i - s_j||$ between them is less than some maximum effective sensing radius $r$, and any such measurement is generated according to:
\begin{equation*}
    \tilde{d}_{ij} = ||s_i - s_j|| + \epsilon_{ij} \qquad    \epsilon_{ij} \sim N(0, \sigma^2)
\end{equation*}
Finally, we assume that there is a subset $A \subset S$ of the sensors (called the \textit{anchors}) whose positions are known exactly \textit{a priori}.

For these experiments,  we use an SNLP instance with 6 estimated nodes and 4 anchors placed on a $6\times6$ unit square in $\mathbb{R}^2$. The maximum range is $r=3$ and the measurements are noiseless but believed to be corrupted by noise $\epsilon_{ij} \sim \mathcal{N}(0, .01)$. The problem is visualized in graphical form in Appendix \ref{snlp-prob-vis}. A ground truth sample containing about 1.1 million points was recovered using a standard nested sampling library, dynesty \cite{dynesty}. Each variational inference method was set to produce a sample of 200 points. Although this problem is relatively small, it is sufficient to reveal significant differences in performance between the different variational inference methods.

\begin{figure*}[t]
  \centering
  \includegraphics[width=.65\linewidth]{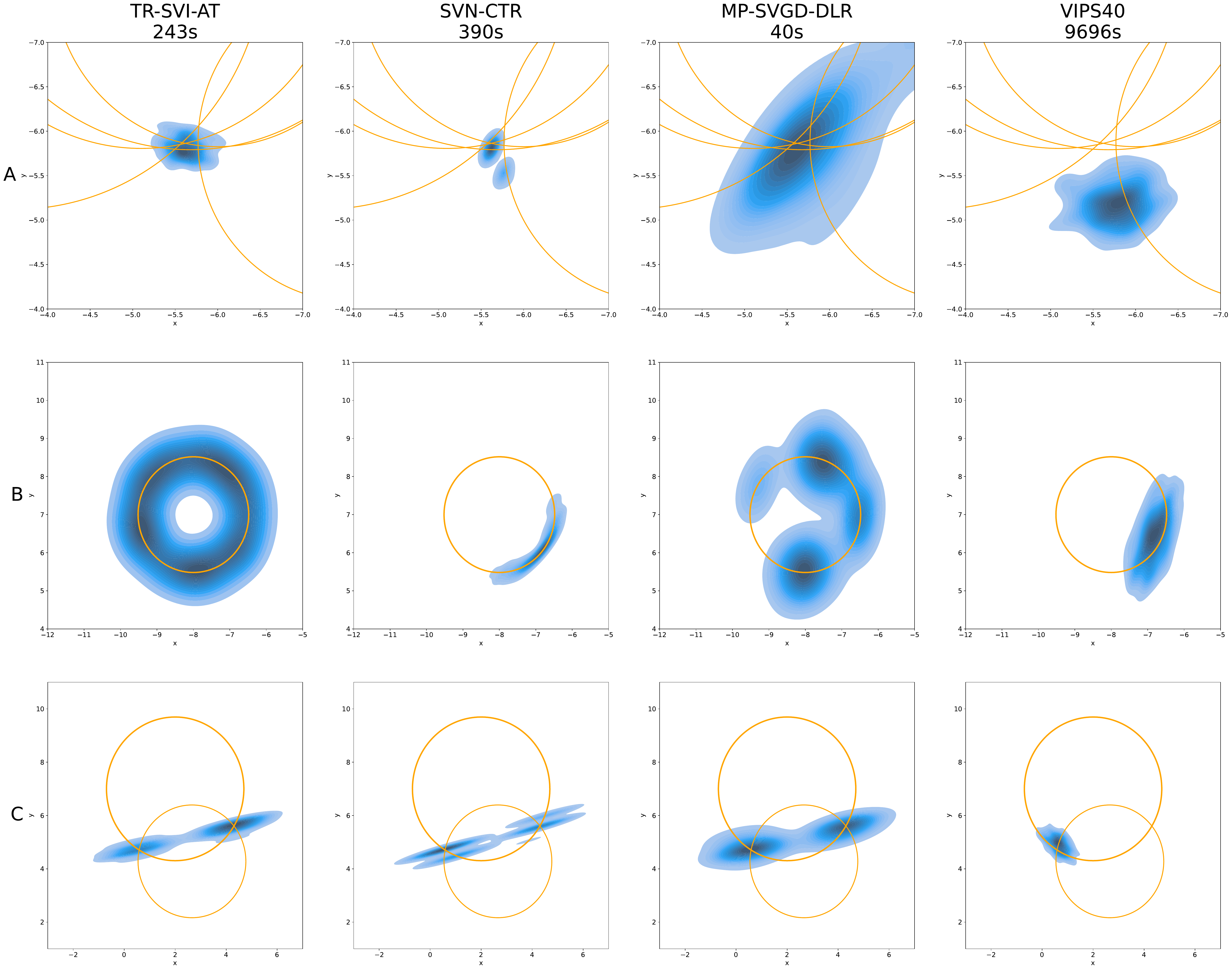}
  \caption{Kernel density estimation (KDE) plots of the final samples produced by various variational inference methods on a high-dimensional, noisy SNLP problem. From each sample, the marginal samples corresponding to the location of a selected sensor are extracted and visualized as a KDE plot. Since ground truth was not recoverable, we also visualize the measurements received by each selected sensor to enable qualitative analysis. These measurements are displayed as orange circles with a radius equal to the range measurement centered on the true position of the sending node. The time to generate the sample (in seconds) is displayed under its name.}
\label{big-qual}
\end{figure*}

\subsubsection{Convergence Results}
Our first experiment in this set focuses on analyzing the convergence behavior of the different SVI methods with different step control approaches. Methods like SVN and MP-SVGD that lack adaptive step control rely on \textit{a priori} user-specified static step sizes or step control rules. These experiments demonstrate the performance gap in convergence behavior between adaptive step size methods and \textit{a priori} step selection rules.

Figure \ref{converge-best}a depicts the convergence for our methods and the modified baselines as a function of iteration number. All the second-order methods converge quickly and smoothly. Both MP-SVGD variants oscillate until the step size decays enough to allow convergence. TR-SVI-KL shows fewer iterations than the other methods because of its step rejection mechanism, which the other methods lack.

Figure \ref{converge-best}b depicts the convergence of the methods as a function of time. Although the per-iteration time complexity of second-order methods is greater than that of first-order methods, they require fewer iterations, and thus less compute time overall, to achieve a small gradient magnitude.

\subsubsection{Numerical Performance of Generated Samples}
Next, we evaluate the accuracy of the generated sample sets as approximations of the target posterior distribution, once again using the MMD metric (Eq. \ref{mmdeq}) against the dynesty reference sample. The results of this experiment are presented in Table \ref{table}.

Our trust-region methods achieved the lowest average MMD scores with the reference sample. VIPS40 performed the worst on this test, potentially due to the difficulty of approximating the annular shapes of the SNLP posterior with a Gaussian mixture. The samples produced by the different methods are assessed qualitatively in Appendix \ref{small-qual-append}.

\subsection{Qualitative Analysis of High-Dimensional SNLP}
Our last set of experiments is intended to test how the different methods scale to a high-dimensional problem with complex shapes. To this end, we apply it to a larger 50 sensor, 12 anchor SNLP instance over a $20 \times 20$ unit square in $\mathbb{R}^2$. For this problem, the maximum sensor range $r=3$ and the measurements received by the sensors are perturbed with noise $\epsilon_{ij} \sim \mathcal{N}(0, .01)$ whose mean and variance are known values. The results of this experiment are only assessed qualitatively since recovering a high-quality reference sample using traditional methods (e.g. dynesty \cite{dynesty}) is intractable. For comparison, MP-SVGD-DLR, SVN-CTR, and VIPS40 were also run on this test. The same set of noisy measurements were used for all methods to ensure a fair comparison.

Figure \ref{big-qual} shows KDE plots generated from the samples produced by each of four methods for three selected sensors and the time required to generate each approximation. Figure \ref{big-qual} also displays the incoming measurements for each sensor as circles to enable easier qualitative analysis. 

Sensor A received multiple measurements, so a unimodal distribution is expected, with some variation due to sensor noise. All SVI methods recovered a unimodal distribution centered correctly on the intersection of the various measurements. VIPS40 recovered a unimodal distribution, but it is not correctly centered. Sensor B received a single range measurement from an anchor, so its posterior should be annular. Of the four methods, only TR-SVI-AT produced a sample with a balanced annular shape. Sensor C received two range measurements, resulting in a bimodal distribution. All SVI methods captured this bimodal distribution but VIPS40 only captured one of the two modes.

Overall, TR-SVI-AT appears to capture intricate details in high dimensions significantly more accurately than previous variational inference methods. Notably, VIPS40 took significantly (\textgreater 20x) longer than the SVI methods and produced a visibly worse approximation. 
\section{Conclusion}

We introduce a SVI method that leverages known conditional independence structure, second-order information, and adaptive step control to ensure good convergence on high-dimensional, non-convex, and ill-conditioned objectives. Our method demonstrated faster and more reliable convergence than existing SVI methods. The approximations produced by our method were more accurate than those of existing SVI methods and a state-of-the-art parametric variational inference method.

\section{Acknowledgements}
Liam Pavlovic was supported by the National Science Foundation Graduate Research Fellowship Program (NSF-GRFP). 
\bibliography{AISTATS2025PaperPack/paper}

%%%%%%%%%%%%%%%%%%%%%%%%%%%%%%%%%%%%%%%%%%%%%%%%%%%%%%%%%%%%
\appendix
\onecolumn
\section{Proof of Equation \ref{mp-grad}} \label{mp-grad-proof}
From the proof of Theorem 3.3 in \cite{svgd}, we have
\begin{equation*}
    \tau\innerproduct{\nabla J[S]}{V} + O(\tau^2) = J[S + \tau V] - J[S]
\end{equation*}
and, at $S=\mathbf{0}$,
\begin{equation*}
 J[\mathbf{0} + \tau V] - J[\mathbf{0}] = -\Delta_1 - \Delta_2
\end{equation*}
where 
\begin{align*}
    &\Delta_1 = \mathbb{E}_{z \sim q} [\log p(z + \tau V(z))] - \mathbb{E}_{z \sim q} [\log p(z)]\\
    &\Delta_2 = \mathbb{E}_{z\sim q} [\log \det(I + \tau \nabla_z V(z))] - \mathbb{E}_{z\sim q} [\log \det(I)]
\end{align*}
For $V \in \mathcal{H}_1 \times ... \times \mathcal{H}_D$ the terms above can be computed as
\begin{align*}
\Delta_1 &= \mathbb{E}_{z \sim q} [\log p(z + \tau V(z))] - \mathbb{E}_{z \sim q} [\log p(z)]\\
&= \tau \mathbb{E}_{z \sim q}[ \nabla_z\log p(z) \cdot V(z)  ] + O(\tau^2) \\
&= \tau \sum_{a=1}^D \mathbb{E}_{z \sim q} [\nabla_{z_a}\log p(z)v_a(z)] + O(\tau^2) \\
&= \tau \sum_{a=1}^D \innerproduct{\mathbb{E}_{z \sim q} [\nabla_{z_a}\log p(z) k_a(z, \cdot)]}{v_a(\cdot)}_{\mathcal{H}_a} + O(\tau^2)
\end{align*}
and 
\begin{align*}
    \Delta_2 &= \mathbb{E}_{z\sim q} [\log \det(I + \tau \nabla_z V(z))] - \mathbb{E}_{z\sim q} [\log \det(I)] \\
    &= \tau\mathbb{E}_{z\sim q} [\text{trace}(I^{-1} \cdot \nabla_z V(z))] + O(\tau^2)\\
    &= \tau \mathbb{E}_{z\sim q} [\text{trace}(\nabla_z V(z))] + O(\tau^2)\\
    &= \tau \sum_{a=1}^D \mathbb{E}_{z \sim q} [\nabla_{z_a} v_a(z)] + O(\tau^2) \\
    &= \tau \sum_{a=1}^D \innerproduct{\mathbb{E}_{z \sim q} [\nabla_{z_a} k_a(z, \cdot)]}{v_a(\cdot)} + O(\tau^2)
\end{align*}
Thus, 
\begin{equation*}
    \innerproduct{\nabla J[\mathbf{0}]}{V} = \sum_{a=1}^D \innerproduct{-\mathbb{E}_{z \sim q} [k_a(z, \cdot)\nabla_{z_a}\log p(z) + \nabla_{z_a} k_a(z, \cdot)]}{v_a}_{\mathcal{H}_a} 
\end{equation*}
and 
\begin{equation*}
    (\nabla J[\mathbf{0}](\cdot))_a = -\mathbb{E}_{z \sim q} [k_a(z, \cdot)\nabla_{z_a}\log p(z) + \nabla_{z_a} k_a(z, \cdot)]
\end{equation*}

\section{Proof of Theorem \ref{second-var-theorem}} \label{second-var-proof}
From the proof of Theroem 1 in SVN \cite{svn}, we know that the second variation of the SVI objective along a pair of directions $V, W \in \mathcal{H}_1\times ... \times \mathcal{H}_D$ equals
\begin{equation}
   D^2J[\mathbf{0}](V, W) = -\mathbb{E}_{z \sim q} \left[ W(z)^\top \nabla^2_{z} \log p(z) V(z) - \text{trace}(\nabla_{z}W(z)\nabla_{z}V(z)) \right]
\end{equation}

By the reproducing properties of $\mathcal{H}_1\times ... \times \mathcal{H}_D$, namely
\begin{equation*}
    v_a(z) = \innerproduct{k_a(z, \cdot)}{v_a(\cdot)}_{\mathcal{H}_a} \qquad w_a(z) = \innerproduct{k_a(z, \cdot)}{w_a(\cdot)}_{\mathcal{H}_a}
\end{equation*}
and
\begin{equation*}
    \partial_{z_b} v_a(z) = \innerproduct{\partial_{z_b} k_a(z, \cdot)}{v_a(\cdot)}_{\mathcal{H}_a} \qquad \partial_{z_b} w_a(z) = \innerproduct{\partial_{z_b} k_a(z, \cdot)}{w_a(\cdot)}_{\mathcal{H}_a}
\end{equation*}
we get
\begin{equation*}
    \mathbb{E}_{z \sim q} \left[ W(z)^\top \nabla^2_{z} \log p(z) V(z) \right] = \sum_{a=1}^D\sum_{b=1}^D \innerproduct{\innerproduct{\mathbb{E}_{z \sim q} \left[ -k_a(z, x)k_b(z,y)\partial^2_{ab} \log p(z)\right]}{w_b(y)}_{\mathcal{H}_b}}{v_a(x)}_{\mathcal{H}_a}
\end{equation*}
 and 
\begin{equation*}
    \mathbb{E}_{z \sim q} \left[ \text{trace}(\nabla_{z}W(z)\nabla_{z}V(z)) \right] = \sum_{a=1}^D\sum_{b=1}^D\innerproduct{\innerproduct{\mathbb{E}_{z \sim q} \left[\partial_{z_a}k_b(z, y)\partial_{z_b}k_a(z, x) \right]}{w_b(y)}_{\mathcal{H}_b}}{v_a(x)}_{\mathcal{H}_a}
\end{equation*}
Plugging these in yields the final expression for the second variation
\begin{equation}
\sum_{a=1}^D\sum_{b=1}^D \innerproduct{\innerproduct{h_{ab}(x, y)}{w_b(y)}_{\mathcal{H}_b}}{v_a(x)}_{\mathcal{H}_a}
\end{equation}
where 
\begin{equation}
     h_{ab}(x, y) = \mathbb{E}_{z \sim q} \left[ -k_a(z, x)k_b(z, y)\partial_{ab} \log p(z) 
     + \partial_{z_a}k_b(z, y)\partial_{z_b}k_a(z, x) \right]
\end{equation}

\section{SNLP Problem Visualizations} \label{snlp-prob-vis}
\begin{figure}[H]
  \centering
  \includegraphics[width=\linewidth]{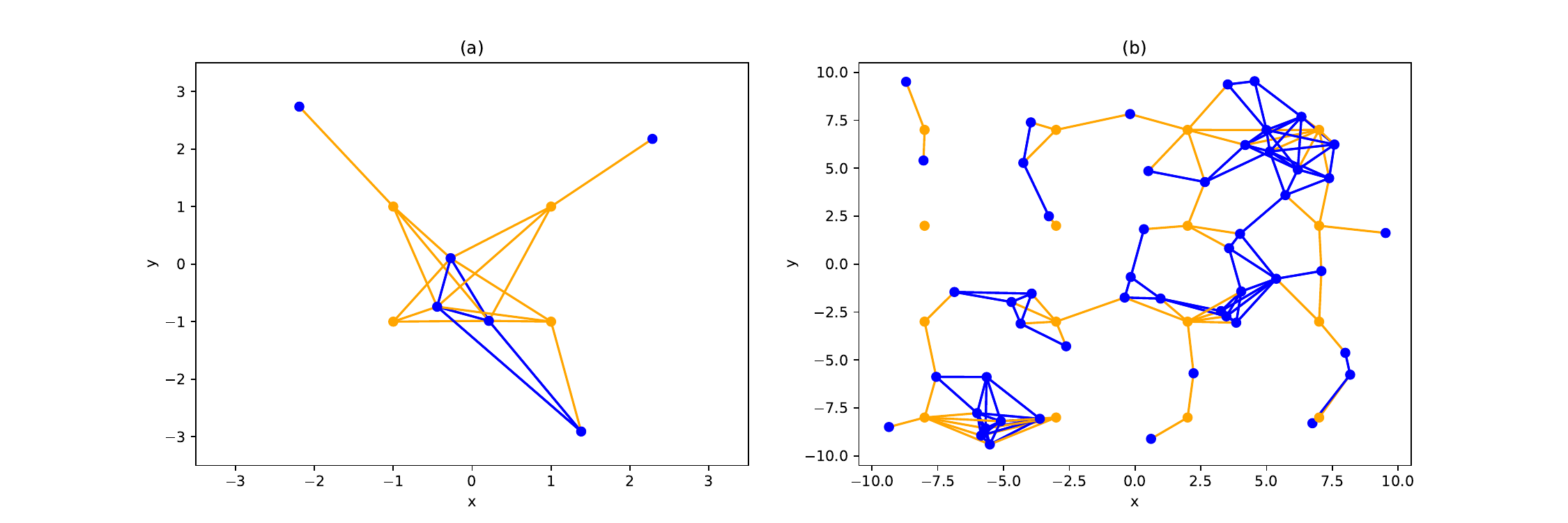}
  \caption{Graph representations of the sensor network localization problems used for evaluation with the small example on the left and the large example on the right. Estimated nodes are depicted in blue and anchors in orange. The edges represent shared range measurements between pairs of nodes. Blue edges correspond to measurements shared between two estimated nodes and orange edges correspond to measurements from an anchor.}
  \label{snlp-graphs}
\end{figure}

\section{Bayes Net Generation Details}\label{bn-generation}

The nodes of these Bayes Nets are organized into layers. The 30-dimensional has 3 layers with 10 nodes each. The 80-dimensional has 4 layers with 20 nodes each. A node $x_j$ from the first layer has a marginally Gaussian distribution $p(x_j) = \mathcal{N}(\mu, \sigma^2)$. A node $x_j$ in any subsequent layer is conditioned on some random $[1, M]$-size subset $C_j$ of the nodes from the previous layer with which it shares connections in the network. The conditional distribution of such a node is either a Gaussian or Gaussian mixture of the form
\begin{equation}
p(x_j | C_j) = \mathcal{N}(\sum_{x_k \in C_j} \alpha_k x_k, \sigma^2) \quad \text{or} \quad p(x_j | C_j) = \sum_{l=1}^2 \omega_l \mathcal{N}(\sum_{x_k \in C_j} \alpha^l_k x_k, \sigma^2)
\end{equation}

where $\{\alpha^{(l)}_k\}$ is a set of weights and $\{\omega_l\}$ are the GMM component weights. The 30-dimensional problem has a total of 6 GMM nodes and the 80-dimensional has 20. Nodes from all layers but the first were uniformly selected at random to be a GMM node. The specific random generative procedure for values of each parameter are
\begin{itemize}
    \item  $\mu$ is selected uniformly from $[0,2]$ for 30-dimensional $[0, 4]$ for 80-dimensional 
    \item All weights $\{\alpha^{(l)}_k\}$ are selected independently and uniformly from $[-1, 1]$
    \item The first GMM weight $\omega_1$ is selected uniformly from $[.4, .6]$, the second $\omega_2$ completes the sum to 1.
    \item All variances $\sigma^2$ were sampled uniformly over orders of magnitude $[10^{-3}, 10^0]$ to induce poor-conditioning
    \item Maximum number of connections $M$ is 3 for 30-dimensional and 4 for 80-dimensional
    
\end{itemize}

\section{Hyperparameter Details} \label{hyperparams}

\begin{table}[h!]
  \caption{Hyperparameter Settings}
  \label{hyperparam table}
  \centering
  \begin{tabular}{lllll}
    \toprule
    Model    & 12-Dim SNLP  & 100-dim SNLP & 30-Dim BN & 80-Dim BN\\
    \midrule
    MP-SVGD-DLR (initial step, step decay) & 0.1, 0.99 & 0.1, 0.99 &  0.01, 0.999 & 0.01, 0.99\\
    MP-SVGD-AG (intial step) & 0.5 & N/A &  0.05 & 0.05 \\
    SVN-CTR   (trust region size)  & 1 & 0.1 &    0.1  & 0.1\\
    Kernel (lengthscale) & 1 & 3 & 10 & 60\\
    
    \bottomrule
  \end{tabular}
\end{table}

The hyperparameters used for each SVI baseline, as well as the kernel hyperparameters utilized by all SVI methods, are shown in Table \ref{hyperparam table}. VIPS40 utilizes the default configuration provided by the source code \cite{vips}. TR-SVI-AT and TR-SVI-KL utilize the parameter settings listed in their respective algorithm blocks (Algorithms \ref{tr-svi-kl}, \ref{tr-svi-at}) for all experiments. The hyperparameters for the trust-region methods were set based on performance on small toy problems during early algorithm development and then used for every experiment without alteration. In general, the convergence of trust-region methods is not sensitive to exact hyperparameter settings \cite{nocedal}. 
 
\section{Additional Convergence Results}\label{add-conv-res}
\begin{figure}[H]
  \centering
  \includegraphics[width=.8\linewidth]{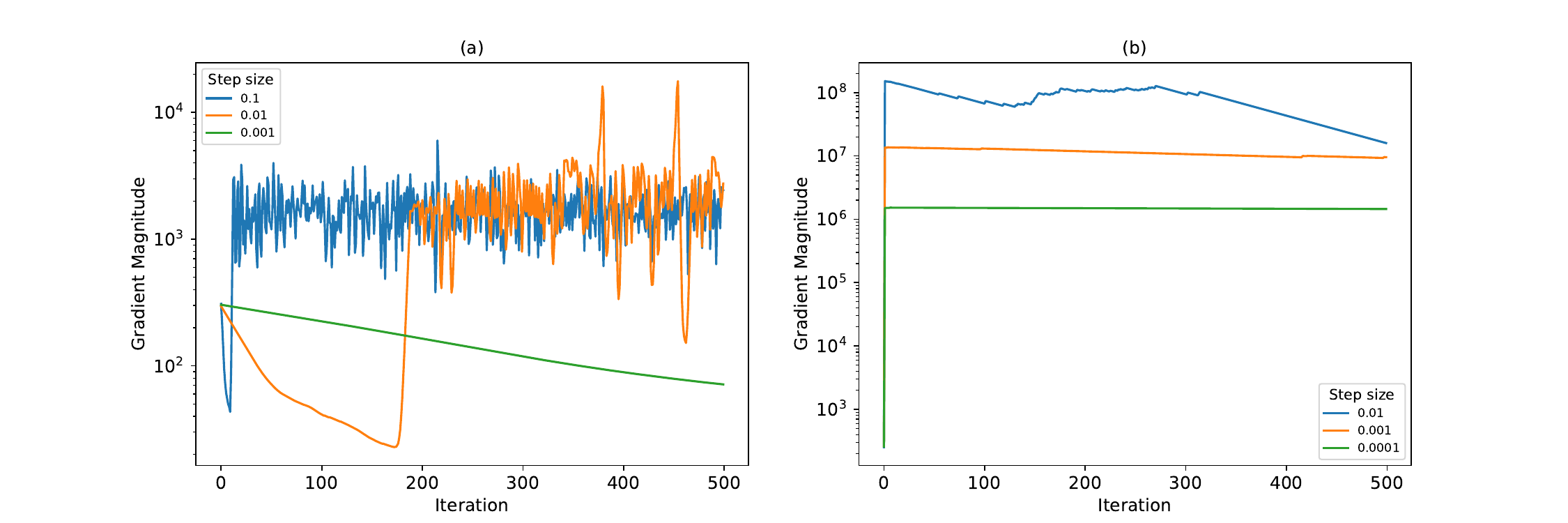}
  \caption{The convergence rates of MP-SVGD(a) and SVN(b) on the small SNLP instance with a variety of static step sizes. None produce good results.}
  \label{together-bad}
\end{figure}

This additional set of tests is designed to demonstrate the necessity of adaptive step control for ensuring convergence on non-convex objectives. To do this, we analyze the convergence of MP-SVGD \cite{mpsvgd} and SVN \cite{svn} on the small SNLP example with static step sizes. Figure \ref{together-bad} depicts the convergence rates of MP-SVGD and SVN with a variety of static step sizes, none of which produce good results. For MP-SVGD, the step size is either too large for a portion of the optimization, causing the method to oscillate over the objective, or too small, resulting in slow convergence. For SVN, all the step sizes result in overly large initial steps from which the method subsequently struggles to recover. The poor performance of these methods in this experiment motivated the introduction of the improved MP-SVGD-DLR, SVN-CTR and MP-SVGD-AG baselines.
\section{Qualitative Analysis of Low-Dimensional SNLP} \label{small-qual-append}
\begin{figure}[H]
  \centering
  \includegraphics[width=.95\linewidth]{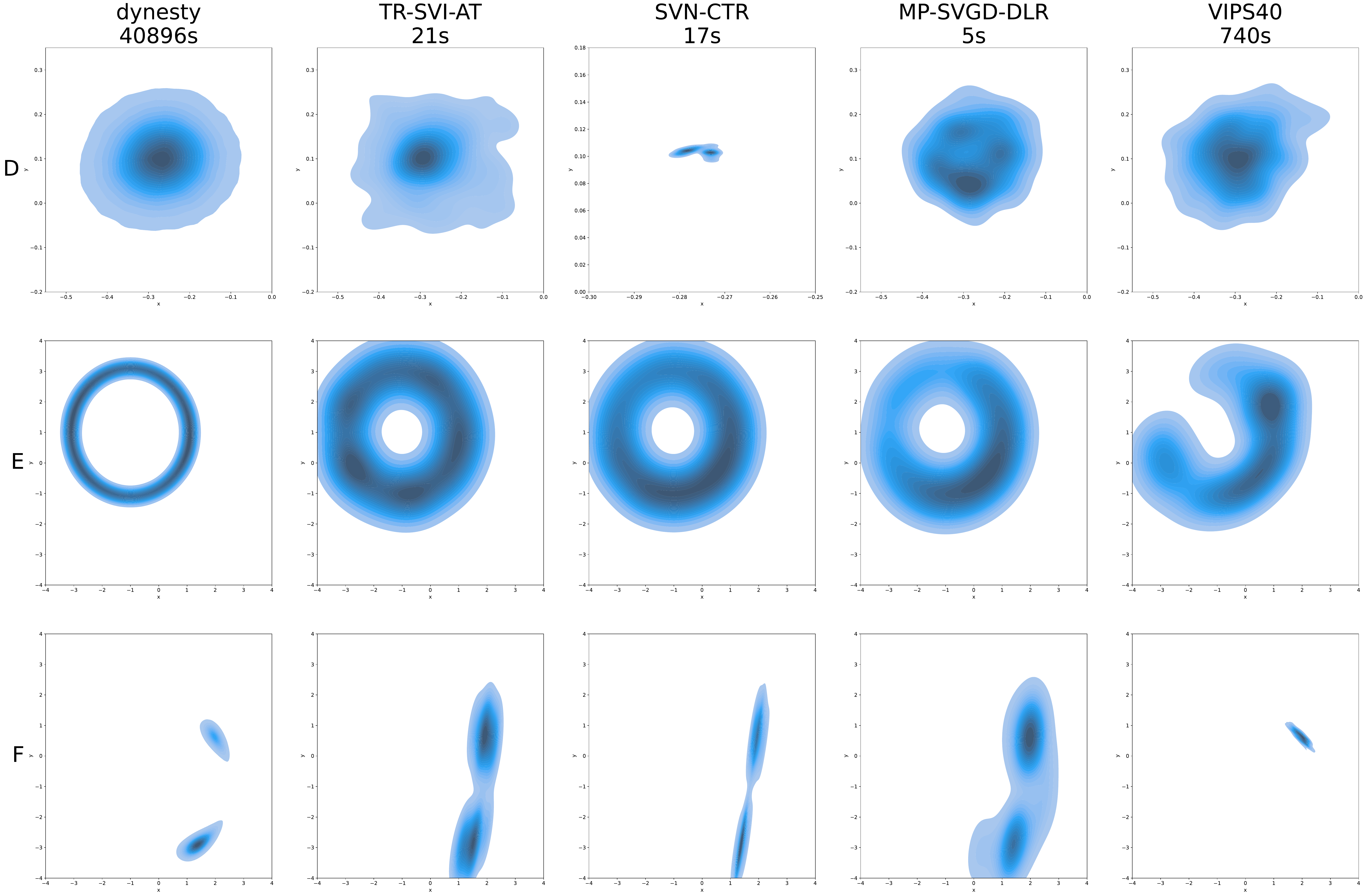}
  \caption{Kernel density estimation (KDE) plots of the final samples produced by various SVI methods and the dynesty reference on a low-dimensional SNLP problem. From each sample, the marginal samples corresponding to the location of the selected sensor are extracted and visualized as a KDE plot. The KDE plots of the different methods are displayed on the same scale with the exception of SVN-CTR's plot for sensor A, which required a scale an order of magnitude smaller to be visible. The time required to generate each sample is displayed under its name.}
\label{small-qual}
\end{figure}

This additional set of tests is designed to demonstrate how differences in MMD performance on the small SNLP instance, as reported in Table \ref{table} translate to perceptible differences in the quality of the different sample approximations. Figure \ref{small-qual} shows kernel density estimate (KDE) plots generated from samples produced by MP-SVGD-DLR, SVN-CTR, TR-SVI-AT, VIPS40, and dynesty, as well as the time required to generate the samples. Plots of the marginal posterior estimates for the locations of three selected sensors are shown. These sensors were selected because they represent a variety of posterior shapes. 

Sensor D received multiple measurements from other nodes, resulting in a dense unimodal distribution. VIPS40, TR-SVI-AT, and MP-SVGD-DLR capture this sensor's posterior well. SVN-CTR, on the other hand, significantly underestimates the variance of the posterior, requiring a different axis scale than the reference distribution to be visible. 

Sensor E received single anchor measurement resulting in a annular posterior. All three SVI methods produced a sample with a annular shape, but the samples of MP-SVGD-DLR and SVN-CTR display a bias towards the bottom right of the annulus not present in TR-SVI-AT's sample. VIPS40 recovers only a partial arc that is misshapen. All 4 methods produce more diffuse distributions than the reference.

Sensor F received a few range measurements but not as many as Sensor D, resulting in a bimodal distribution. All 3 SVI methods capture the bimodality of the distribution, but they also assign more probability to the lower probability mode than the reference distribution. VIPS40 does not capture the bimodality of this distribution, only capturing the lower probability mode.

\end{document}